# Semantic Nutrition Estimation: Predicting Food Healthfulness from Text Descriptions

Dayne R. Freudenberg[1], Daniel G. Haughian[1], Mitchell A. Klusty, BS[1], Caroline N. Leach, BS[1], W. Scott Black, MD[2], Leslie N. Woltenberg, PhD[2], Rowan Hallock[1], Elizabeth Solie[1], Emily B. Collier, MS[1], Samuel E. Armstrong, MS[1], V. K. Cody Bumgardner, PhD[1]

[1]Center For Applied Artificial Intelligence, University of Kentucky, Lexington, KY
[2]College of Health Sciences, University of Kentucky, Lexington, KY

**Abstract**

*Accurate nutritional assessment is critical for public health[1], but existing profiling systems require detailed data often unavailable or inaccessible from colloquial text descriptions of food. This paper presents a machine learning pipeline that predicts the comprehensive Food Compass Score 2.0[2] (FCS) from text descriptions. Our approach uses multi-headed neural networks to process hybrid feature vectors that combine semantic text embeddings, lexical patterns, and domain heuristics, alongside USDA Food and Nutrient Database for Dietary Studies[3] (FNDDS) data. The networks estimate the nutrient and food components necessary for the FCS algorithm. The system demonstrated strong predictive power, achieving a median $R^2$ of 0.81 for individual nutrients. The predicted FCS correlated strongly with published values (Pearson's r = 0.77), with a mean absolute difference of 14.0 points. While errors were largest for ambiguous or processed foods, this methodology translates language into actionable nutritional information, enabling scalable dietary assessment for consumer applications and research.*

**Introduction**

Accurate nutritional profiling of foods is essential for informing public health policy[1], guiding consumer choices[4], and enabling large-scale dietary analysis[5]. However, evaluating the nutritional value of food items is a complex task, often requiring detailed nutritional knowledge and access to comprehensive food composition data. For the average consumer, or even large-scale research efforts, translating a list of nutrients into an actionable understanding of a food's overall quality can be overwhelming[6]. The Food Compass scoring system addresses this challenge by reducing multidimensional nutritional information into a single, interpretable Food Compass 2.0 score. Scores range from 0 to 100, with higher values indicating healthier options. The system evaluates food across nine domains: nutrient ratios, vitamins, minerals, food ingredients, additives, processing, specific lipids, fiber and protein, and phytochemicals. Scores in each domain are summed to create a raw FCS. The raw score is then clipped to an empirically determined range based on 5th and 95th percentile of scores in the Food and Nutrient Database for Dietary Studies (FNDDS) dataset (-12.80 to 29.42), as outlined in the Food Compass methodology, and linearly transformed to create the final FCS on the 1 to 100 scale. With this calculation transformed into an intuitive metric, individuals are empowered to make informed dietary decisions.

While the FCS offers a valuable tool for simplifying nutritional assessment, the Food Compass 2.0 system itself has acknowledged limitations. For example, the system's rating of foods based on a 100-calorie portion can make direct comparisons between certain foods confusing. The distinction between processed and unprocessed foods, while improved, remains limited. Furthermore, the system does not fully align with environmental or sustainability measures, and questions persist about how well it captures the overall nutritional quality of specific foods, particularly those rich in certain vitamins, minerals, or from animal sources[7][8][9].

Although the Food Compass offers a simplified measure of overall food healthfulness, it remains somewhat inaccessible because scoring requires extensive nutrient data and understanding of the underlying algorithm. Scoring a food item requires an accurate and comprehensive nutritional profile. Obtaining these values is especially burdensome for mixed or multi-component foods, such as complex or composite meals, or colloquially described dishes. Because these informal or aggregated descriptions often lack sufficient detail[10] on individual ingredients and preparation methods, reliably calculating their nutritional content is a major obstacle to integrating nutrient profiling systems like the Food Compass algorithm for practical, user-facing applications. In this work, we present a machine learning pipeline that enables automatic nutritional estimation from plain-text food descriptions, removing the need for manual nutrient input while enhancing the accessibility and broadening the applications of the FCS profiling system. The pipeline begins by converting food text descriptions into embeddings, which are numerical vectors that encode linguistic information. These embeddings position words or phrases with similar meanings close to one another in a high-dimensional space, capturing semantic relationships, and enabling the model to understand nuance and context.

For instance, in this space, the vector for "grilled chicken" would be mathematically closer to "roast chicken" than to "chocolate cake." Our work created a custom embedding model on the USDA Food and Nutrient Database for Dietary Studies (FNDDS), Food Patterns Equivalents Database[12] (FPED), and Database for the Flavonoid Content of Selected Foods[13] to extract rich, semantic representations of food descriptions. These datasets have micronutrient and macronutrient amounts, food component parts, and specific flavonoid information, respectively, for a given set of food descriptions. For each neural network, the first head converted unstructured food descriptions into semantically meaningful numeric representations, while the second head compared these representations to generate a complete nutrient profile.

Once the nutritional profile of a food is estimated, the results are passed through the Food Compass algorithm to generate a comprehensive healthfulness score, complete with a breakdown of the underlying nutritional factors that contributed to that score. This creates a seamless pipeline that translates everyday food descriptions, such as "grilled chicken sandwich with lettuce and tomato," into a meaningful health score that is both accessible to non-experts and actionable in research, public health, or consumer settings.

By combining natural language processing with structured nutritional modeling, our system makes it possible to perform scalable, real-time dietary evaluations from unstructured inputs. This capability has wide-reaching implications, from powering diet-tracking and meal-logging applications to enabling automated assessments of food environments, restaurant menus, or grocery inventories. The goal in this work is to aid in nutritional decision-making at every level. We offer a new approach for democratizing access to healthful eating guidance, bringing the rigor of nutritional science into everyday decision-making through the simplicity of the Food Compass Score.

**Methods**

The workflow to predict and score the required nutrient information of the Food Compass 2.0 scoring system is comprised of four stages: (1) creating a numerical representation of food descriptions by combining text embeddings, word frequency information, specific cooking processes, and domain knowledge of ingredient components; (2) training a multi-headed neural network per nutrient described in the FCS methodology to simultaneously synthesize a single, comprehensive food-specific embedding and predict a nutrient value; (3) aggregating these values into a Food Compass 2.0 score; and (4) validating the pipeline against published Food Compass 2.0 scores.

**Creating Numerical Representations of Food Descriptions**

FNDDS datasets for 54 target nutrients and food attributes were obtained. Food entries with irregular data (e.g. N/A and negative values) were removed. A rule-based filtering mechanism was used, which automatically detected non-food-related terms (e.g., "truck", "computer") to maintain training data quality. We created a data augmentation pipeline to increase the training dataset, therefore deepening the relationship between food descriptions and numerical nutrient values. Using a random set (30%) of food descriptions, we generated additional food descriptions by applying context-sensitive synonym replacement (e.g., replacing "chicken" as "poultry"), including common cooking varieties (e.g., "grilled" or "baked"), and by adding in portion size qualifiers (e.g., "large" and "small") to existing food descriptions. This was accomplished through keyword searches and random assignments. The 30% sample was decided as a large enough sample to establish model-relevant correlations without decreasing the models' generalization capabilities. The associated nutrient value of these augmented samples was adjusted based on heuristic weighted multipliers (e.g. 1.5x calories when adding "fried" to an existing food description). Using heuristics was optimal, as it reinforced existing food description-nutrient value relationships without needing precise multiplier value assignments. The augmented descriptions were combined with the unaltered descriptions for training. The models trained with augmentation had an increased $R^2$ score when generalizing to unseen data than training an identical model without augmentation.

We built an aggregate feature vector to effectively summarize complex subtleties present in food text descriptions, combining data from three different sources. First, to capture semantic connotations, we used the pre-trained Sentence Transformer model all-MiniLM-L6-v2[14] to convert each food description into a dense 384-dimension vector. The all-miniLM-L6-v2 was selected for its ability to remain scalable and computationally efficient, without a substantial loss in performance. Second, to represent lexical regularities and keyword importance, we chose to apply a Term Frequency-Inverse Document Frequency[15] (TF-IDF) vectorizer, which was set to output a maximum size vector of 1024 elements based on individual terms and two-word sequences (n-grams), excluding common English stop words. TF-IDF establishes relationships between common words and nutritional values by enumerating every word in the training data and assigning value based on word frequency counts. Last, we assembled a set of 15 to 20 nutrient-specific heuristic features that inject expert knowledge of nutritional value in relation to keywords in the input text. These features included numerical indicators for different cooking methods (e.g., 1.5 for 'fried'), binary features for terms indicating high or low nutrient content (e.g., "fortified," "sugar-free"), and text statistics such as numbers of words and characters. We decided to include the heuristic features to assist the models in understanding the entire context of and establishing relationships between different input strings. The sets of features were then combined in order to generate an input vector for the neural networks, which was then standardized using a Standard Scaler to increase model performance.

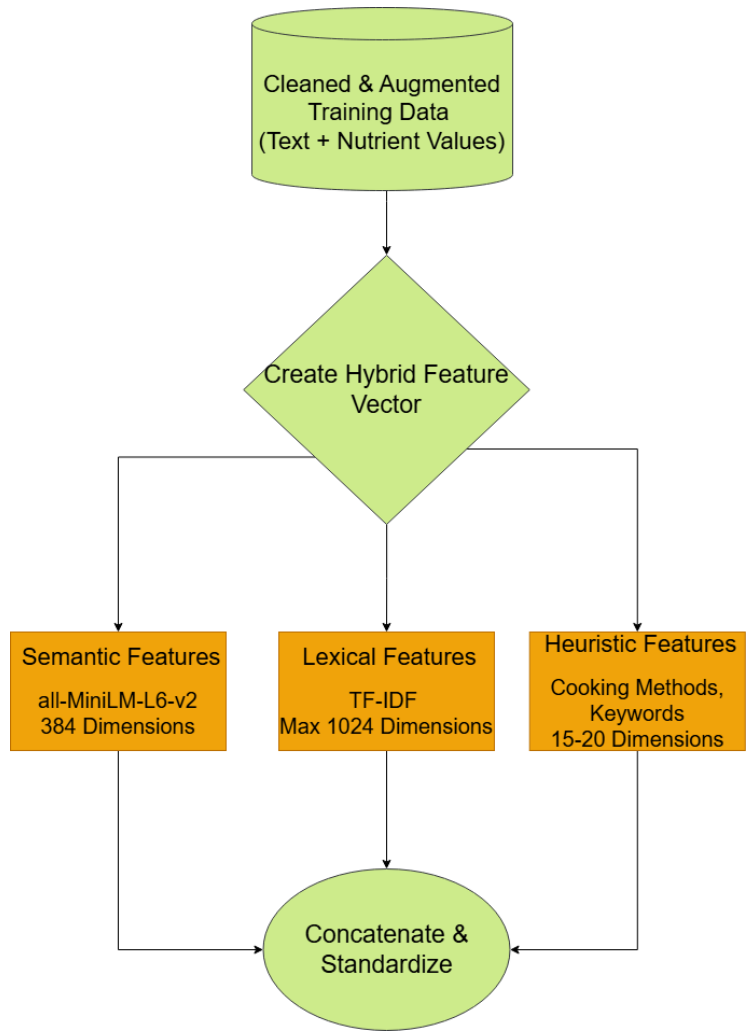

*Figure 1:* Hybrid feature vector creation

**Neural Network Training**

Neural networks are computational models adept at learning patterns from complex, non-linear, high-dimensional data like that of our hybrid feature vector. We chose this architecture for its multi-use capability in simultaneously creating a custom embedding vector and predicting the nutrients accordingly. Our goal was to use a two-step neural network designed to learn food-specific embeddings to estimate nutrients that can then be applied to the scoring procedure outlined in the Food Compass 2.0 methodology. The network, in the first instance, processed the high-dimension hybrid feature vector through a feature encoder. The feature encoder consisted of several fully connected linear layers (Input -> 1024 -> 768 -> Embedding Dimension) separated by ReLU[16] activation functions and dropout layers (p=0.3) to handle overfitting. The embedding dimension was adjusted to fit each nutrient model, commonly 1024, 512, 256, and 128.

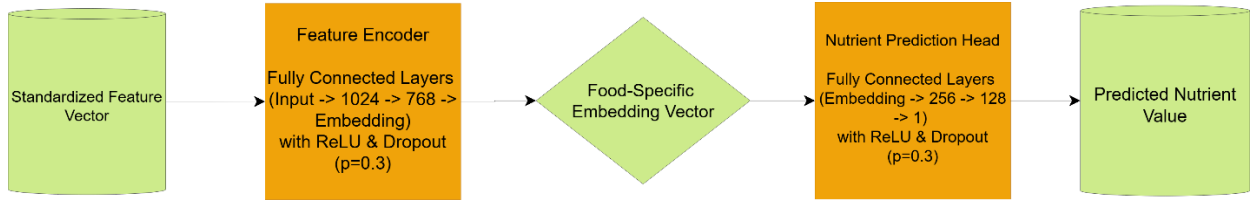

*Figure 2: Neural network architecture*

The embedding was subsequently transferred to a nutrient prediction head. A distinct prediction head, typically structured with three linear layers (e.g., Embedding Dimension -> 256 -> 128 -> 1), was specifically trained for each of the nutrients in the Food Compass 2.0 scoring system domains. For each target, the model underwent training utilizing a Mean Squared Error (MSE) loss function and a batch size of 32 to penalize the model for making significant errors. A learning rate scheduler (ReduceLROnPlateau[17]) featuring a patience of 8 epochs alongside Adam[18] optimizer (initial learning rate = 5e-4) was implemented to facilitate dynamic adjustments to the learning rate after a relatively short time, if model performance was not increasing significantly. We utilized early stopping with a patience of 15 epochs to promote model generalization and ease of training, resulting in the halting of training when validation loss failed to improve. These hyperparameters were tuned at the time of training. The efficacy of each model was assessed on a reserved set of training data, employing the coefficient of determination ($R^2$), Root Mean Squared Error (RMSE), and Mean Absolute Error (MAE) as standard evaluation metrics.

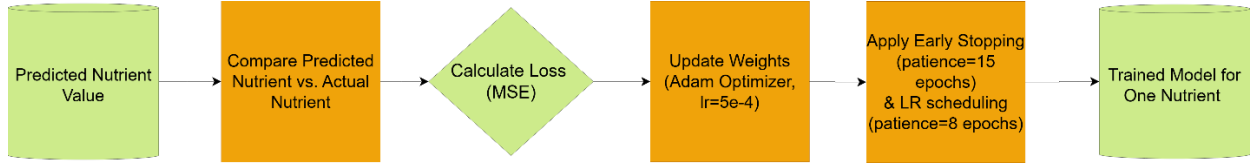

*Figure 3: Neural network training loop*

**Food Compass Scoring**

Our adapted Food Compass 2.0 scoring process involved multiple modifications from the original scoring system due to the programmatic challenges involved in processing unstructured input text. For instance, in the Processing domain, we replaced the existing NOVA scoring to allow for non-integer NOVA category numbers using linear interpolation. We interpolated the points as outlined in the scoring procedure: (1, 10), (2, 7.5), (3, 5), (4, -10). Additionally, we used keyword-based heuristics to score fermentation and frying. Similarly, we omitted the five binary additive attributes as well as dried vegetables and fruits' scores. These heuristic approaches and omissions were judged as the most optimal choices for programmatically implementing the Food Compass 2.0 algorithm.

The algorithm is built on a primary scoring function that scales a given predicted nutrient value ($v$) to a score between a minimum ($p_{min}$) and maximum ($p_{max}$), after clipping it $(l)$ $(h)$ range.

$$S(v, l, h, p_{min}, p_{max}) = p_{min} + (p_{max} - p_{min}) \cdot \frac{clip(v, l, h) - l}{h - l}$$

The nutrient ratio domain ($D_1$) scores the logarithmic balance between key nutrients. Each ratio is calculated only if a specific check is met. First, fat quality is calculated only if energy from fat is below 10 percent. The log-ratio of unsaturated ($U$) to saturated fat ($S$) is scored on a -10 to 10 scale, halved for dairy products:

$$S_{fat} = S\left(\ln\left(\frac{U_g}{S_g}\right), -0.66, 1.77, -10, 10\right)$$

Second, carbohydrate quality is calculated only if energy from carbs is less than $10(F)$. The log-ratio of fiber ($C$) to total carbohydrates ($C$) is scored on a -10 to 10 scale:

$$S_{carb} = S\left(\ln\left(\frac{F_g}{C_g}\right), -7.02, -0.78, -10, 10\right)$$

Third, the potassium sodium ratio is calculated only if potassium is less than 10 mg and sodium is less than 10 mg (per 100 kcal). The log-ratio of potassium ($K$) to sodium ($Na$) is scored on a -10 to 10 scale:

$$S_{KNa} = S\left(\ln\left(\frac{K_{mg}}{Na_{mg}}\right), -2.02, 3.30, -10, 10\right)$$

Scores are calculated for 12 vitamins in the vitamin domain ($D_2$). The scoring function parameters are listed in the table below. The final domain score is the average of the top 5 individual scores by absolute value.

| Vitamin | $l$ | $h$ | $p_{min}$ | $p_{max}$ |
|---|---|---|---|---|
| Vitamin A, RAE | 0 mcg | 225.0 mcg | 0 | 10 |
| Niacin (B3) | 0 mg | 4.0 mg | 0 | 10 |
| Vitamin C | 0 mg | 22.5 mg | 0 | 10 |
| Vitamin B6 | 0 mg | 0.325 mg | 0 | 10 |
| Vitamin D (D2+D3) | 0 mcg | 3.75 mcg | 0 | 10 |
| Folate, DFE | 0 mcg | 100 mcg | 0 | 10 |
| Vitamin E | 0 mg | 3.75 mg | 0 | 10 |
| Vitamin B12 | 0 mcg | 0.6 mcg | 0 | 10 |
| Vitamin K | 0 mcg | 30.0 mcg | 0 | 10 |
| Choline | 0 mg | 137.5 mg | 0 | 10 |
| Thiamin (B1) | 0 mg | 0.3 mg | 0 | 10 |
| Riboflavin (B2) | 0 mg | 0.325 mg | 0 | 10 |

Table 1: *Vitamin domain scoring procedure*

Scores are calculated for 9 minerals in the mineral domain ($D_3$). The scoring function parameters are listed in the table below. The final domain score is the average of the top 5 individual scores by absolute value:

| Mineral | $l$ | $h$ | $p_{min}$ | $p_{max}$ |
|---|---|---|---|---|
| Calcium | 0 mg | 250.0 mg | 0 | 10 |
| Iron | 0 mg | 4.5 mg | 0 | 10 |
| Magnesium | 0 mg | 105.0 mg | 0 | 10 |
| Phosphorus | 0 mg | 175.0 mg | 0 | 10 |
| Potassium | 0 mg | 1175.0 mg | 0 | 10 |
| Zinc | 0 mg | 2.75 mg | 0 | 10 |
| Copper | 0 mg | 0.225 mg | 0 | 10 |
| Selenium | 0 mcg | 13.75 mcg | 0 | 10 |
| Sodium | 0 mg | 575.0 mg | -10 | 0 |

Table 2: *Mineral domain scoring procedure*

The score for the ingredients domain ($D_4$) is the sum of points from beneficial and harmful ingredients:

| Ingredient | l (cup/oz. eq.) | h (cup/oz eq.) | $p_{min}$ | $p_{max}$ |
|---|---|---|---|---|
| Fruits | 0 | 1.75 | 0 | 10 |
| Non-starchy vegetables | 0 | 4.77 | 0 | 10 |
| Beans & legumes | 0 | 0.50 | 0 | 10 |
| Nuts & seeds | 0 | 1.35 | 0 | 10 |
| Whole grains | 0 | 1.12 | 0 | 10 |
| Seafood | 0 | 3.86 | 0 | 10 |
| Yogurt | 0 | 0.81 | 0 | 10 |
| Plant oils (grams) | 0 g | 11.31 g | 0 | 10 |
| Refined carbohydrates | 0 | 1.38 | -10 | 0 |
| Red & processed meat | 0 | 2.69 | -10 | 0 |

*Table 3: Ingredient domain scoring procedure*

The additive domain score ($D_5$) is the average of two sub-scores. The first is added sugar, which is scored with a stepwise function based on the percentage of calories from added sugar ($P_{sugar}$). This score ranges from 0 points (for $P_{sugar} = 0$) down to -10 points (for $P_{sugar} = 60$). The second is nitrite meat, scored based on the percentage of calories from cured meat ($P_{nitrate}$):

$$S_{nitrate} = S(P_{nitrate}, 0, 50, -10, 0)$$

The processing domain score ($D_6$) combines three factors. These are the NOVA score (a linear interpolation: NOVA 1 is +10, 2 is +7.5, 3 is +5, and 4 is -10), a fermentation score (scored positively based on fermented content percentage), and a frying score (-10 if the food is fried, otherwise 0). The final score is a weighted combination:

$$D_6 = \frac{S_{NOVA} + (0.5 \cdot S_{fermentation}) + (0.5 \cdot S_{frying})}{2}$$

Scores are calculated for the four lipids detailed in the specific lipids domain ($D_7$). The final score is a weighted average of the top 3 scores, with the result multiplied by 0.5:

| Lipid | l | h | $p_{min}$ | $p_{max}$ | Weight |
|---|---|---|---|---|---|
| Cholesterol | 0 mg | 75 mg | -10 | 0 | 0.5 |
| EPA+DHA | 0 g | 0.0625 g | 0 | 10 | 1.0 |
| ALA | 0 g | 0.4 g | 0 | 10 | 0.5 |
| MCT | 0 g | 0.32 g | 0 | 10 | 0.5 |

*Table 4: Specific lipid domain scoring procedure*

The fiber and protein domain score is $S_{fiber} = S(fiber_g, 0, 9.5, 0, 10)$ combines fiber and protein, with protein given a half-weight. The individual scores are $S_{protein} = S(protein_g, 0, 14, 0, 10)$ and $S_{protein} = S(protein_g, 0, 14, 0, 10)$. The final domain score is:

$$D_8 = \frac{S_{fiber} + (0.5 \cdot S_{protein})}{1.5}$$

Last, the phytochemicals domain averages points for two phytochemicals, and the entire domain is given a half-weight: $S_{flavonoids} = S(flavonoids_{mg}, 0, 23.53, 0, 10)$, $S_{carotenoids} = S(carotenoids_{mcg}, 0, 8746.81, 0, 10)$:

$$D_9 = 0.5 \cdot \frac{S_{flavonoids} + S_{carotenoids}}{2}$$

The nine individual domain scores $(D_i)$ are then summed and clipped to form the final FCS $(S_{final})$.

$$S_{final} = \text{round}\left(100 - \frac{99 \times \left(29.94 - \text{clip}(\sum_{i=1}^{9} D_i, -12.43, 29.94)\right)}{42.37}\right)$$

**Prediction Validation**

To measure our system's overall accuracy, we set up a validation pipeline to compare our predicted score against the published Food Compass 2.0 scores based on the corresponding FNDDS nutritional data. To begin, a plain text description matching a published Food Compass 2.0 score was fed into our prediction pipeline, which generated the total nutritional profile unique to that description. The resulting predicted outputs were, in turn, assessed by our adapted FCS algorithm to generate a Food Compass Score 2.0. The predicted score was then compared against its corresponding published Food Compass 2.0 score. We calculated an absolute difference in scores to quantify prediction errors. This comparison measured the model's overall effectiveness by calculating mean and median absolute difference and identifying categories of food showing the most difference from scores in the reference dataset.

**Results**

*Performance of Individual Nutrient Prediction Models*

The performance of the 48 individual nutrient and food component prediction models was evaluated on a held-out test set. Predictive accuracy for the NOVA classification model was 0.957. The coefficient of determination ($R^2$) for key models is summarized in Table 1.

| Ingredients | R² | Vitamins | R² | Minerals | R² |
|---|---|---|---|---|---|
| Seafood | 0.976 | Vitamin K | 0.916 | Magnesium | 0.873 |
| Cured Meat | 0.894 | Choline | 0.908 | Calcium | 0.865 |
| Red Meat | 0.887 | Thiamin | 0.838 | Selenium | 0.852 |
| Beans & Legumes | 0.896 | Vitamin B6 | 0.830 | Potassium | 0.804 |
| Nuts & Seeds | 0.858 | Vitamin A | 0.829 | Iron | 0.799 |
| Grains (total) | 0.822 | Riboflavin | 0.801 | Phosphorus | 0.751 |
| Refined Grains | 0.854 | Niacin | 0.784 | Zinc | 0.751 |
| Fruits | 0.825 | Vitamin E | 0.793 | Copper | 0.706 |
| Non-starchy Vegetables | 0.830 | Folate | 0.787 | Sodium | 0.697 |
| Yogurt | 0.714 | Vitamin B12 | 0.776 | | |
| Plant Oils | 0.740 | Vitamin C | 0.722 | | |
| Added Sugar | 0.870 | Vitamin D | 0.709 | | |

*Table 5: Performance of Ingredient, Vitamin, and Mineral Models*

| Lipids | R² | Phytochemicals | R² | Nutrients | R² |
|---|---|---|---|---|---|
| DHA (22:6 n-3) | 0.920 | Carotenoids (total) | 0.829 | Calories | 0.884 |
| EPA (20:5 n-3) | 0.802 | Flavonoids | 0.021 | Protein | 0.858 |
| Cholesterol | 0.907 | | | Carbohydrates | 0.847 |
| Capric Acid (10:0) | 0.767 | | | Unsaturated Fat (total) | 0.833 |
| ALA (18:3) | 0.753 | | | Saturated Fat | 0.732 |
| Lauric Acid (12:0) | 0.725 | | | Fiber | 0.74 |
| Caprylic Acid (8:0) | 0.444 | | | | |

*Table 6:* Performance of Lipid, Phytochemical, and Nutrient Models

The models demonstrated strong overall predictive power with a median R² of 0.81 across all targets. As shown in Tables 5 and 6, the models' performance depended heavily on whether a component was explicitly mentioned in the text. For example, predictions for ingredients like "seafood" were highly accurate because, for most seafood items and dishes, the seafood component is referenced directly in the description. In contrast, performance was poor for components like flavonoids, which struggle to be inferred from name alone and must be extracted from a defined ingredient list. It is unlikely that the poor performance of the flavonoid model significantly contributes to the error in the Food Compass 2.0 score prediction, as its domain is given half weight for scoring. Tables 5 and 6 summarizing the R² for all 48 models are found above.

*Accuracy of Food Compass Score 2.0 Prediction*

The performance of the complete prediction pipeline was evaluated against the published Food Compass 2.0 scores for 9,241 food items. Our predicted Food Compass 2.0 scores demonstrated a strong correlation with the published Food Compass 2.0 scores, achieving a Pearson correlation coefficient of r = 0.77 (p < 0.001). This statistic quantifies the strength of the linear relationship between our predicted scores and the actual scores. Our result indicates a strong positive association, meaning that as the actual score for a food item increases, our predicted score also consistently increases, and the p-value confirms this relationship is statistically significant. The overall error was quantified with a Mean Absolute Difference (MAD) of 14.0 points and a Median Absolute Difference of 11.0 points on the 1-100 FCS scale. The distribution of predicted scores (Mean = 43.4, SD = 25.0) closely tracked the actual scores (Mean = 47.3, SD = 27.9), though a slight systematic underprediction was observed (Mean difference = 3.96). Analysis of accuracy thresholds showed that 64.6% of predictions fell within ±15 points of the actual score, and 84.0% of predictions fell within ±25 points.

Analysis of prediction errors revealed that discrepancies were not uniformly distributed across food types. The largest errors were concentrated in two main areas: complex, multi-ingredient processed foods and single-ingredient foods with nutritionally vague, ambiguous, or incomplete descriptions. For processed foods, categories with the highest MAD included "Mixed grain-encased dishes" (e.g., pizza, burritos) and "Mixed meat dishes". For instance, "Baked Alaska" was predicted to have a score of 84.0 while its actual score is 3.0, an error of 81.0 points. In these cases, the embedding model encoding cannot accurately extract model-relevant details due to the lack of information the descriptions hold. For single-ingredient foods, large errors occurred when the description was ambiguous or had conflicting nutritional characteristics accounted for in the Food Compass 2.0 algorithm. "Mushroom soup, canned, undiluted" (Actual: 3.0, Predicted: 100.0) was likely misinterpreted by the model, which focused on "mushroom" without adequately penalizing for "canned soup," a highly processed format. Similarly, "Cranberries, NS as to raw, cooked, or canned" (Actual: 14.0, Predicted: 100.0) highlights a failure case where the lack of processing information leads to an overly optimistic prediction. Conversely, the model sometimes failed to recognize the healthfulness of less common foods, such as "Chamnamul, cooked, fat added in cooking" (Actual: 99.0, Predicted: 22.0). To account for these shortcomings, more detailed descriptions are needed in training to increase the generalization capabilities of the neural network and embedding models. In addition, user-defined ingredient specifications could increase performance with existing models without the need for retraining. Each point in Figure 3 represents one of the 9,241 food items from the validation dataset. The strong clustering of points along y = x indicates a high degree of correlation and low systematic bias in the prediction pipeline.

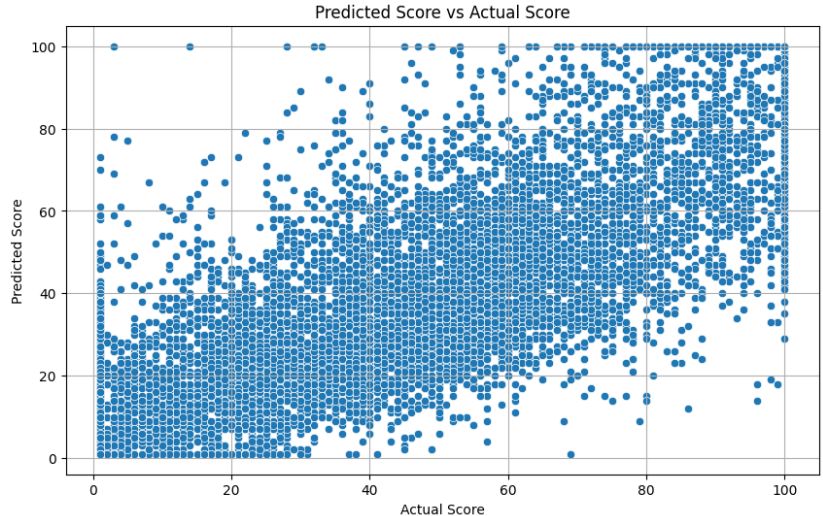

*Figure 4: Correlation between predicted and actual Food Compass Scores*

**Discussion and Conclusions**

This study demonstrates that machine learning can successfully bridge the gap between text-based food descriptions and quantitative nutritional assessment. The strong predictive performance of our FCS score prediction pipeline validates a proof of concept for generating a healthfulness metric from text alone. We achieved a Pearson correlation coefficient ($r = 0.77$, $p < 0.001$) for overall FCS predictions and a median $R^2$ of 0.81 for individual nutrient estimates. This validation removes the primary accessibility barrier: requiring detailed, pre-existing nutrient data to evaluate a food's nutritional quality. As a result, large scale, automated nutritional analysis of food given text descriptions is feasible, enabling new applications in public health and consumer-facing technology.

The primary limitation of our approach is its difficulty in capturing the nutritional complexity of highly processed foods from sparse textual information. The overall MAD of 14.0 points is predominantly driven by errors in two categories: complex multi-ingredient processed foods and single-ingredient items with ambiguous descriptions. Descriptions for items like "Pizza" or "Canned Soup" lack the critical details that influence nutritional value, leading to higher prediction errors. For example, the item "Baked Alaska" exhibited an 81.0-point MAD compared to the published score. And in the case of "Mushroom soup, canned, undiluted", processing details were not adequately penalized. The informational deficit is further exemplified by the extremely low predictive accuracy for components that must be inferred without explicit textual cues, such as flavonoids ($R^2 = 0.021$). This highlights an inherent challenge: the semantic gap between a simple food-name and its variable composition. This can be addressed by incorporating more advanced natural language processing models capable of inferring processing details, by developing an interactive system that prompts users for clarification when ambiguity is detected, such as a Large Language Model or by utilizing a food-oriented foundational model with inherent domain knowledge.

In conclusion, our work provides a methodology for text-based nutritional prediction using a hybrid feature vector approach to overcome the data accessibility bottleneck inherent in traditional FCS calculation. By translating natural language into health scores, this research lays out a foundation for a new generation of scalable, accessible dietary assessment tools that can be deployed in real-world settings while acknowledging the dependency on input specificity necessary to mitigate significant prediction errors in highly complex or underspecified food items.